\documentclass[sn-mathphys-num]{sn-jnl}

\usepackage{graphicx}%
\usepackage{multirow}%
\usepackage{amsmath,amssymb,amsfonts}%
\usepackage{amsthm}%
\usepackage{mathrsfs}%
\usepackage[title]{appendix}%
\usepackage{xcolor}%
\usepackage{textcomp}%
\usepackage{manyfoot}%
\usepackage{booktabs}%
\usepackage{algorithm}%
\usepackage{algorithmicx}%
\usepackage{algpseudocode}%
\usepackage{listings}%
\usepackage{bm}
\usepackage{svg}

\usepackage{float}
\usepackage{algorithm}
\usepackage{amssymb}
\usepackage{pifont}
\newcommand{\cmark}{\ding{51}}%
\newcommand{\xmark}{\ding{55}}%

\theoremstyle{thmstyleone}%
\theoremstyle{thmstyletwo}%

\theoremstyle{thmstylethree}%

\begin{document}

\title[Embedding spatial context in urban traffic forecasting with contrastive pre-training]{Embedding spatial context in urban traffic forecasting with contrastive pre-training}


\author*[1]{\fnm{Matthew} \sur{Low}}\email{mattchrlw@gmail.com}

\author[1]{\fnm{Arian} \sur{Prabowo}}\email{arian.prabowo@unsw.edu.au}

\author[1]{\fnm{Hao} \sur{Xue}}\email{hao.xue1@unsw.edu.au}

\author[1]{\fnm{Flora} \sur{Salim}}\email{flora.salim@unsw.edu.au}

\affil*[1]{\orgdiv{School of Computer Science and Engineering}, \orgname{UNSW Sydney}, \orgaddress{\street{Engineering Rd}, \city{Kensington}, \postcode{2033}, \state{NSW}, \country{Australia}}}


\abstract{Urban traffic forecasting is a commonly encountered problem, with wide-ranging applications in fields such as urban planning, civil engineering and transport. In this paper, we study the enhancement of traffic forecasting with \textit{pre-training}, focusing on spatio-temporal graph methods. While various machine learning methods to solve traffic forecasting problems have been explored and extensively studied, there is a gap of a more contextual approach: studying how relevant non-traffic data can improve prediction performance on traffic forecasting problems. We call this data \textit{spatial context}. We introduce a novel method of combining road and traffic information through the notion of a \textit{traffic quotient graph}, a quotient graph formed from road geometry and traffic sensors. We also define a way to encode this relationship in the form of a \textit{geometric encoder}, pre-trained using contrastive learning methods and enhanced with OpenStreetMap data. We introduce and discuss ways to integrate this geometric encoder with existing graph neural network (GNN)-based traffic forecasting models, using a contrastive pre-training paradigm. We demonstrate the potential for this hybrid model to improve generalisation and performance with zero additional traffic data. Code for this paper is available at \url{https://github.com/mattchrlw/forecasting-on-new-roads}. \textit{This research/project was undertaken with the assistance of resources and services from the National Computational Infrastructure (NCI), which is supported by the Australian Government. This work was also supported by the NCI National AI Flagship scheme, with computational resources provided by NCI Australia, an NCRIS enabled capability supported by the Australian Government.}}


\maketitle

\section{Introduction}\label{ch:intro}

\begin{figure}[H]
    \centering
    \includegraphics[width=\linewidth]{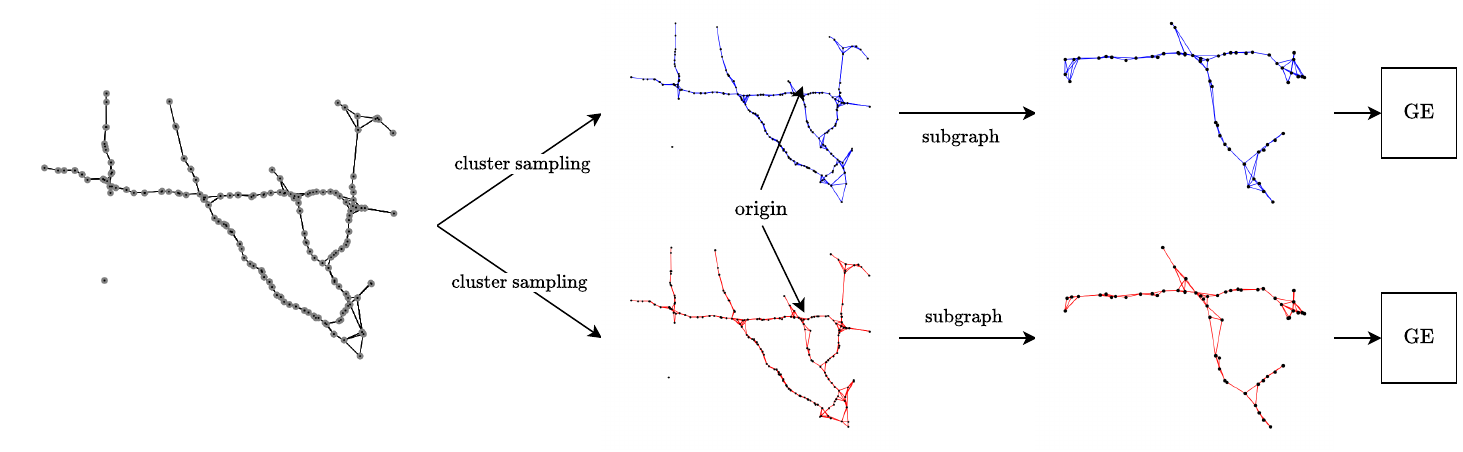}
    \caption{Our geometric encoder pre-training method constructs \textit{traffic quotient graphs}, generates stochastic pairs of graphs augmented with real OpenStreetMap contextual data, and feeds them into a geometric encoder. This figure illustrates this process applied to the METR-LA dataset.}
    \label{fig:enter-label}
\end{figure}

Graphs are a powerful tool for modelling and problem solving, as they map easily to many systems in our world. As each vertex (or edge) of a graph can be associated with data, they can be used to represent diverse datasets from biology to transportation, and are commonly used in commercial applications as wide-ranging as Google Maps and Amazon recommendations \cite{Veli_kovi__2023}. 

In certain contexts, vertices have additional meaning as they represent some sort of geographic position; we call this data \textit{spatial graph data}.\footnote{Usually, the actual position is encoded as a feature on the vertex.} If spatial graph data is \textit{observed}, or otherwise measured at different times, this results in data known as \textit{spatio-temporal graphs}.
With a spatio-temporal graph, there are a number of questions one might want to ask. In this paper, we concern ourselves particularly with the problem of \textit{forecasting}; predicting the future value of some variable given the past and present information. Importantly, the graph structure of our spatio-temporal data allows us to—informally—use the proximity and ``connection'' of points (or nodes) to other points when forecasting into the future.

While spatio-temporal forecasting problems can be solved with more traditional, statistical time-series methods, we are particularly interested in approaching the problem with \textit{machine learning methods}, which are trained on existing spatio-temporal data. However, spatio-temporal forecasting is complex, and often the machine learning models used to do so are complex too. To alleviate this complexity, it might make sense to instead train a model on a different, smaller problem and then use that knowledge for the larger, more difficult problem. This machine learning technique is known as \textit{pre-training}.

In this paper, we focus on \textit{urban traffic data}, which can be represented as a graph, with each node representing a traffic sensor measuring some aspect of traffic (such as traffic speed or car density). While there are existing pre-training methods that use existing traffic data to inform future traffic predictions, we wish to find a new (or extend an existing) pre-training technique that applies non-traffic data to improve traffic prediction. In particular, \textit{spatial context}---information around the traffic sensors---could help inform future predictions and improve generalisation ability. This would be one of the first instances in the literature of integrating non-traffic data for solving traffic forecasting problems. In summary, our research questions are as follows:

\begin{enumerate}
    \item How can pre-training paradigms be applied to the spatio-temporal forecasting setting, in particular to improve performance, generalisation and pattern recognition? 
    \item Can we better understand road patterns in spatio-temporal graph-based traffic forecasting by integrating spatial contextual information with traffic sensor information? 
    \item Can we improve performance of forecasting by doing so? 
\end{enumerate}

\begin{figure}
    \centering
    \includegraphics[width=\linewidth]{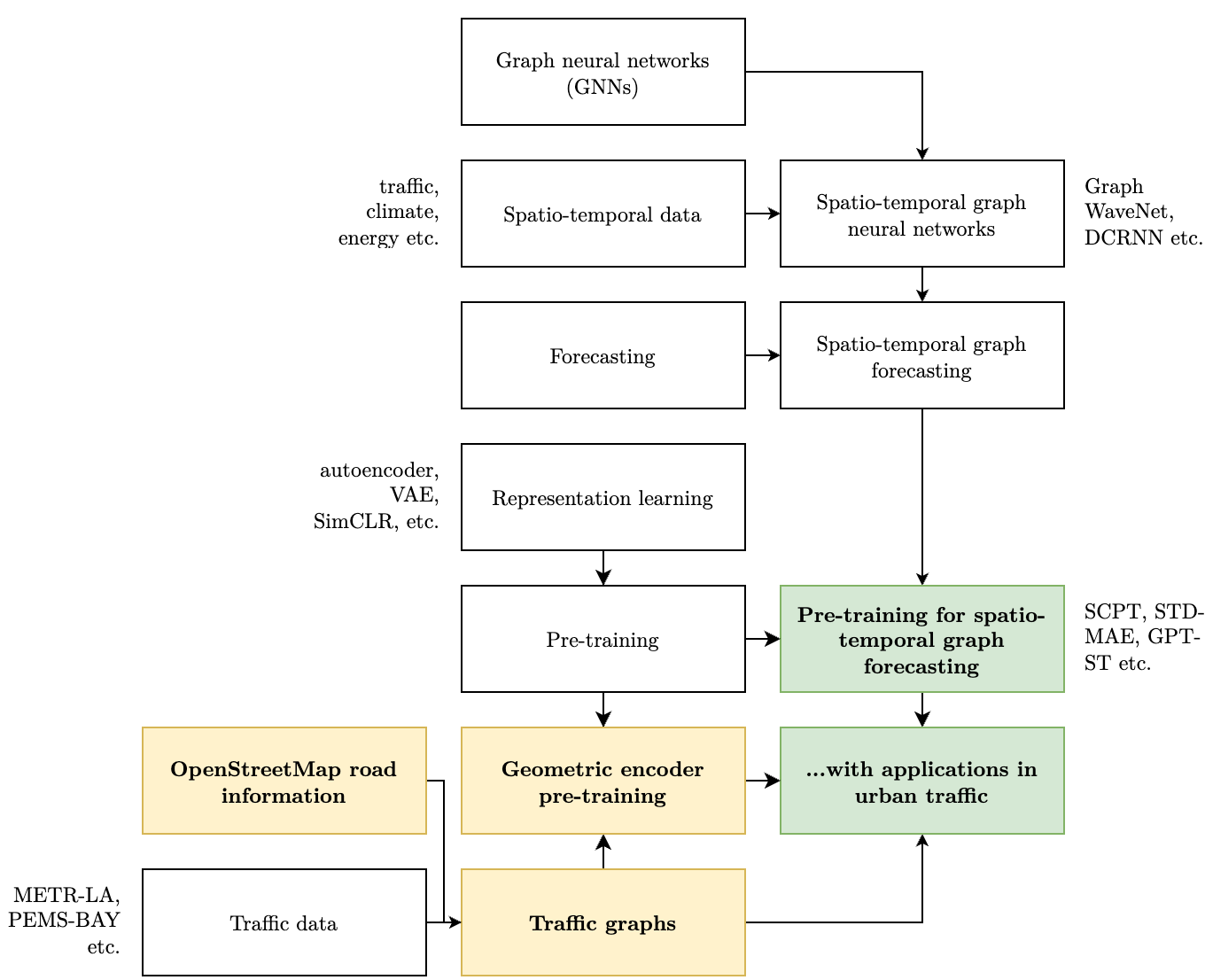}
    \caption{A diagram of the various components of the paper, with the contribution areas highlighted in yellow and the application area in green.}
    \label{fig:enter-label}
\end{figure}

\section{Background}

\subsection{Graph data and spatio-temporal data}

A graph $G = (V, E)$ is a mathematical structure consisting of vertices (or nodes) $V$ and edges $E$ connecting pairs of vertices from $V$. These edges can have directions (directed) or be bi-directional links between vertices (undirected), and these edges can also have weights representing the degree that vertices are linked (weighted) or be binary (unweighted). A \textit{spatio-temporal graph} refers to a graph associated with spatio-temporal data; \textit{one} way to express this is with a pair $[\bm X^{(t)}, G]$ where $\bm X^{(t)}$ refers to a feature matrix at time $t$ associated with a graph $G$. Note that in this definition, $G$ is static throughout the timesteps $t$, although in some cases $G$ can change.

\subsection{Graph neural networks}

\subsubsection{Traditional neural networks}

Deep learning models based on neural networks are the basis for the methods we will discuss in this paper. There are a number of deep learning paradigms or subfields that are of interest to us in this section. In particular, \textit{convolutional neural networks} (CNNs) (a certain type of \textit{feedforward neural network}) have gained popularity for their ability to learn patterns from visual or two-dimensional data. \textit{Recurrent neural networks} (RNNs) propagate information in a bi-directional manner, allowing for a certain level of memory to be embedded to neural networks, and resulting in its increased use in language modeling and recognition applications.

\subsubsection{Graph neural network mechanisms}

One of the fundamental questions that early researchers faced was how to reconcile methods commonly used on images, such as CNNs, with the more abstract type of a graph, where the limitations on structure such as 2D neighbourhoods no longer apply. A \textit{graph neural network} (GNN) is a neural network that accepts this graph structure. The common thread among almost all GNNs is the utilisation of \textit{information propagation} or \textit{message passing}, a technique which involves updating each node's representations based on information gathered from neighbouring nodes.

\subsubsection{Graph convolutional networks}

Most graph neural networks are built on the concept of a \textit{graph convolution}. This is an operation applied to a graph, akin to a \textit{convolutional filter} (think a 2-dimensional sliding window) in a convolutional neural network, but with some differences to handle the graphical nature of the data and the variable neighbourhood sizes of graphs.

Following the notation of Kipf and Welling \cite{kipf2017semisupervisedclassificationgraphconvolutional}, a \textit{graph convolutional layer} has the propagation rule\[
    H^{(l+1)} = \sigma (\tilde D^{-\frac12} \tilde A \tilde D^{-\frac12} H^{(l)} W^{(l)}),
\]
where $\tilde A$ is an adjacency matrix with self-loops, $\tilde D_{ij} = \sum_j \tilde A_{ij}$ is a diagonal degree matrix and $W^{(l)}$ is a trainable weight matrix for the layer. $\sigma$ is an activation function (such as ReLU or sigmoid) and $H^{(l)}$ is the activation matrix in the $l$th layer, with $H^{(0)} = X$. More generally, we can define a graph convolution network, a neural network consisting of graph convolution layers, as \[
\bm Z = f(X, \bm A) = \tilde{\bm A} \bm X \bm W,
\]
where $\bm Z$ is the output, $\tilde{\bm A}$ is the normalised adjacency matrix with self-loops, $\bm X$ is the input signals and $\bm W$ is the model parameter matrix.

\subsection{Spatio-temporal deep learning}

Now that we have introduced basic machine learning methods with graphs, we will begin with a brief overview of deep learning methods designed for the analysis and modelling of \textit{spatio-temporal graph data}. This will help us understand how our modifications fit into existing architectures. Most of these deep learning methods consist of two components: a spatial component for modelling movement between nodes in the graph, and a temporal component for handling flows of information as time passes. 

There are many spatio-temporal methods in the literature; we refer readers to \cite{Jiang_2021} for a more comprehensive survey. However, as our research question focuses on \textit{augmenting} existing forecasting methods with pre-training, we will focus on one specific method ubiquitous in the research community as a benchmark; Graph WaveNet \cite{ijcai2019p264}. This method:
\begin{itemize}
    \item includes a \textit{self-adaptive adjacency matrix} that allows the model to infer hidden dependencies during training time, and;
    \item uses a convolutional neural network for the temporal component, but also incorporates\textit{dilation} of the gated causal convolution layers. This allows the model to scale its receptive field exponentially instead of linearly with respect to the number of layers, simplifying the model architecture.
\end{itemize}

The final structure of the network in the paper consists of $K$ \textit{spatio-temporal layers}, which each consist of two gated TCNs (temporal convolution networks) and one GCN, followed by a linear projection to output. The number of spatio-temporal layers dictates the number of temporal levels. To train this model, we use \textit{mean absolute error} (MAE), defined by \[
L_{\bm \Theta} (\hat{\bm X}^{(t+1):(t+T)}) = \frac{1}{TND} \sum_{i=1}^T \sum_{j=1}^N \sum_{k=1}^{D} \left| \hat{\bm X}_{jk}^{(t+i)} - \bm X_{jk}^{(t+i)}\right|,
\] where $T$ is the range of time-steps, $N$ is the number of nodes and $D$ is the number of features.

\subsection{Spatio-temporal pre-training}

\subsubsection{Spatial contrastive pre-training and spatially gated addition}

Pre-training, in its most general form, is a paradigm for training neural networks which involves training on a problem $A$ and then `applying' that training (weight initialisation etc.) to problem $B$ in a second phase known as \textit{fine-tuning}. Spatial contrastive pre-training (SCPT) \cite{Prabowo_2023} is one of the first examples of applying pre-training to the spatio-temporal prediction setting. In this paper, a \textit{spatial encoder} module is introduced and trained at pre-training time, and then integrated into the main prediction model using a \textit{spatially gated addition} layer. The novelty of this approach is that it allows unseen edges (roads, in this case) to be ``discovered'' and added to the topology of the graph during inference time. This spatially gated addition will form the basis of our proposed spatio-temporal pre-training method.

\subsubsection{Other spatio-temporal pre-training methods}

Briefly we will discuss some other spatio-temporal pre-training methods, starting with \textit{STEP} \cite{Shao_2022}, or \textit{STGNN is Enhanced by a scalable time series Pre-training model} (where STGNN refers to \textit{spatio-temporal graph neural networks}). This paper is motivated by the masked auto-encoder model called \textit{TSFormer}, and performs masking on the input time series, with some masking ratio $r$ ($r$ is set to 75\% in the paper). The encoder operates on unmasked patches, and the decoder operates on the full set of patches. While the masking can be effective to discover long-term temporal patterns, this method does not perform masking on the spatial axis.

\textit{STD-MAE} \cite{ijcai2024p442}, which stands for \textit{Spatio-temporal decoupled masked auto-encoder}, splits up the data into spatial and temporal components, and mask each of the components respectively. Those temporal and spatial representations are then fed into a downstream spatio-temporal predictor. This paper differs from STEP as it not only performs masking on the temporal axis, but on the spatial axis too. 

The \textit{GPT-ST} framework introduced in \cite{li2023gptst} follows a similar pre-training architecture, but introduces a hypergraph to encode patterns between \textit{regions} rather than individual nodes. Specifically, a \textit{hypergraph capsule clustering network} is introduced. A hypergraph is of the form $H = (V, E, \bm H)$, where $V$ is a set of vertices, $E$ is a set of $H$ hyperedges, which each connect multiple vertices, and $\bm H \in \mathbb R^{N \times H}$, which represent the vertex-hyperedge connections. $\bm H$ is learnable in this architecture. While this paper provides an efficient and novel method, with results close to state-of-the-art on many datasets, we wish to investigate simpler methods.

Notably, all of the above methods operate on the same data to pre-train the model as training the model. In the next section, and in the rest of the paper, we will discuss how we plan to use a \textit{different} type of data to pre-train an existing model, with one application in mind: traffic.

\section{Traffic forecasting}

Traffic forecasting has gained notoriety as one of the most common ``toy problems'' for spatio-temporal graph models, and has been the subject of a number of surveys \cite{Jiang_2021}. Consider the structure of a traffic network as a graph $G = (V, E)$, with the vertices $v \in V$ being road sensors, and their values being some measurement of traffic (either speed, traffic flow, or some other variable; it can be multiple variables as well), and the edges $e \in E$ being road connections between those sensors. These can be roads, or some sequence of roads eventually connecting two sensors, and can be weighted or unweighted.

Formally, the road connections can be represented using an adjacency matrix $\bm A \in \mathbb R^{N \times N}$, where $N = |V|$ (the number of vertices). At each time $t$, there is a feature matrix $\bm X^{(t)} \in \mathbb R^{N \times D}$, where $D$ is the number of features. For simplicity, we take $D = 1$ and let $D$ represent speed. We wish to learn some \textit{prediction function} $f$, relating the past data to the future data. In the formulation from \cite{ijcai2019p264}, this is expressed as \[
[\bm X^{(t - S):t}, G] \overset{f}{\to} \bm X^{(t+1):(t+T)},
\]
where $S$ is the number of steps in the past, and $T$ is the number of steps in the future. Note that in this formulation, $G$ is static, but we will discuss methods where the adjacency \textit{adapts} to the graph as time passes.\\

\subsection{Traffic datasets}

The most common benchmark datasets for traffic forecasting\footnote{Note: there are also datasets for \textit{grid-based} traffic forecasting, based in cities with more grid-like road topologies (Manhattan). We focus on graph-based traffic forecasting.} are the METR-LA dataset, based on the Los Angeles metropolitan area in the United States, and PEMS-BAY dataset, based in the South Bay region of the San Francisco Bay area (also in the US). We also include the PEMS-D7(m) dataset, a similar dataset to PEMS-BAY but with data collection limited to weekday periods. Due to the ``road topologies'' of Los Angeles and San Francisco, with their mix of grid-like and more sporadic road patterns, they are useful baseline datasets. It is particularly important to note that these datasets mainly contain traffic sensors on \textit{major highway nodes}; other datasets with smaller roads warrant further investigation, however there is a lack of existing datasets in this area \cite{Jiang_2021}.

\begin{table}
    \small
    \centering
    \begin{tabular}{l|lcccc}
        Name & Region & Nodes & Edges & Time range & Delta\\\hline
        METR-LA & US (LA) & 207 & 1515 & 2012-03-01 -- 2012-06-27 & 5 mins \\
        PEMS-BAY & US (SF) & 325 & 2694 & 2017-01-01 -- 2017-06-30 & 5 mins\\
        PEMS-D7(m) & US (SF) & 228 & 7304 & 2012-05-01 -- 2012-06-30 & 5 mins
    \end{tabular}
    \caption{A comparison of three popular datasets commonly applied to spatio-temporal graph forecasting. Note that the PEMS-D7(m) dataset consists of only weekday data.}
    \label{tab:my_label}
\end{table}

Each node represents a \textit{traffic sensor}, which measures the average speed (as previously discussed, $D = 1$) of vehicles passing through the sensor over a timestep. As provided in the datasets, each edge represents proximity to another sensor, as measured by a \textit{radial basis function} (RBF)\footnote{A radial basis function, often denoted $\varphi$, is one that only depends on its distance between the ``origin'' and itself, so $\varphi(\bm x) = \varphi(\| \bm x\|)$.}, and an edge exists between two nodes merely if the nodes are close together. Notably, this is not the true connectivity of the roads. Later, we will discuss ways we updated the adjacency matrix to better reflect true road geometry, by combining information about the road edges with the locations of the traffic sensors.

\begin{figure}[H]
\centering
\includegraphics[width=\textwidth]{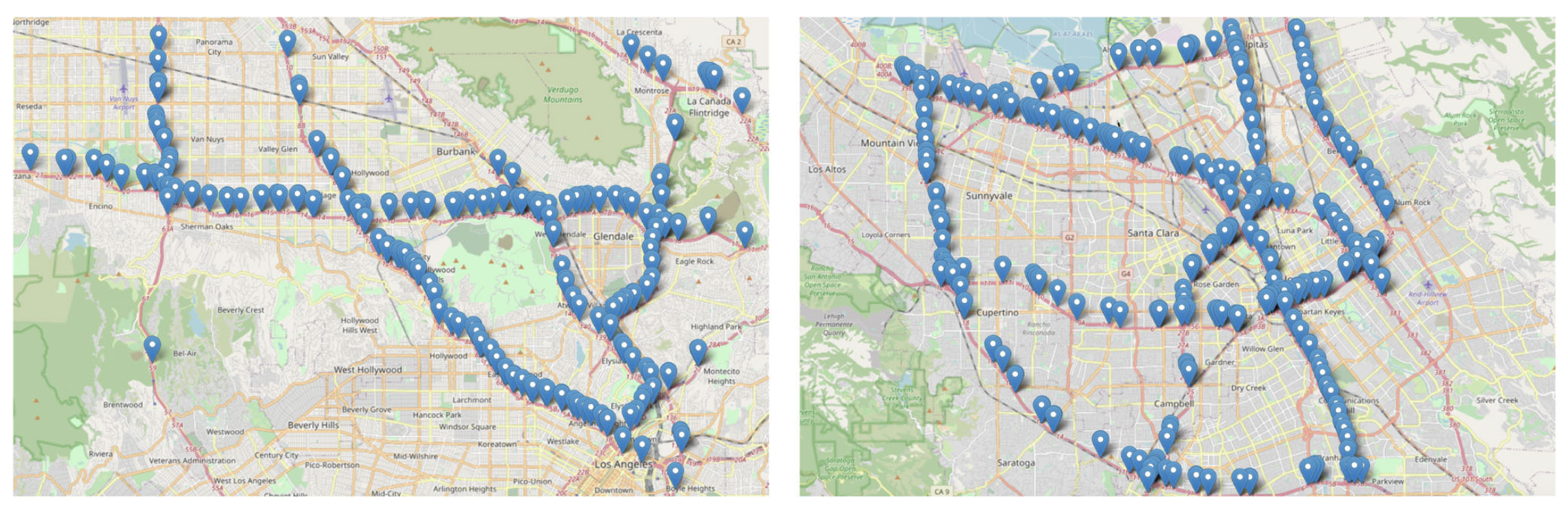}
\caption{The METR-LA and PEMS-BAY datasets \cite{10.1007/s10489-021-02648-0}.}
\end{figure}

\subsection{Pre-training in traffic forecasting}

All of the previously mentioned spatio-temporal pre-training methods used traffic forecasting as one of their toy problems in their papers; a table summarising these methods is below.

\begin{table}
    \centering
    \begin{tabular}{c|c|l|l|l}
        Work & Year & Venue & Method & Data\\\hline
        STEP & 2022  & SIGKDD 2022 & temporal masking & traffic \\
        STD-MAE & 2023 & IJCAI-2024 & spatial and temporal masking & traffic \\
        GPT-ST & 2023 & NeurIPS 2023 & hypergraph clustering & traffic\\
        SCPT & 2023  & ECML PKDD 2023 & contrastive learning & traffic \\
        ? & 2024 & ? & contrastive learning & non-traffic
    \end{tabular}
    \caption{A comparison of spatio-temporal pre-training methods in the literature. Note the lack of methods working with non-traffic data.}
    \label{tab:my_label}
\end{table}

\subsection{The research gap: spatial context}

As we have discussed, there are existing pre-training methods such as SCPT \cite{Prabowo_2023} that aim to improve the performance of prediction of future traffic with existing traffic data. However, there is a distinct gap of methods that improve performance of traffic forecasting by apply pre-training with non-traffic data. In particular, one subject of interest is the information \textit{around} traffic sensors. There is a wealth of information about the areas around traffic sensors---which we call \textit{spatial context}---which is freely available through platforms such as OpenStreetMap, and does not require additional data collection like regular traffic does. This information can have direct impact on the traffic; for example, if a traffic node has a particularly high density of amenities near it, it is likely going to be both a source and target of traffic flow at certain times of the day (such as commuting hours or night-time activity periods).

In this paper, we aim to cover this research gap by proposing methods to combine information such as connectivity, speed limits, road length, points of interest (PoI) and other contextual information around traffic nodes, and integrate these features into the model to allow for learning of traffic around \textit{neighbourhoods}. If these features are shared across different distant traffic nodes in a dataset, then pre-training the dataset with features from existing nodes would allow for better forecasting of traffic nodes which share characteristics in their neighbourhoods but might geographically be far apart. Most promising is that any methods of these can be applied to \textit{all roads, worldwide}.

In the next few sections, we outline the following contributions towards our originally proposed research questions:
\begin{enumerate}
    \item We introduce the notion of a \textit{traffic quotient graph}, and apply this to Graph WaveNet as a way to better understand traffic structure.
    \item We obtain \textit{spatial context} through queries to OpenStreetMap.
    \item We propose a \textit{geometric encoder} and use spatially gated addition to integrate this information into existing spatio-temporal graph forecasting models.
\end{enumerate}

\section{Constructing traffic graphs}

Our first contribution in this paper is related to how we map road information into traffic. The eventual graph of traffic sensor nodes connected and containing information about their surroundings is what we call a \textit{traffic quotient graph}. First, we will define what is needed to construct a traffic quotient graph, and then devise an algorithm doing so, using the concept of \textit{quotient graphs}.

\subsection{Traffic data and road graphs}

Recall that we can formulate traffic data as a graph. To avoid confusion between other graphs that we will discuss, we formulate the traffic graph $T = (T_V, T_E)$ as a graph such that $V$ is the locations of each traffic sensor is stored as pairs in $T_V$, and $T_E = \{\}$ (i.e. there are no edges between traffic sensors. This is not to be confused with a \textit{traffic quotient graph} which we will soon define.

A \textit{road graph} consists of road nodes connected with road edges. These represent the real geometry of roads. For instance, if a road contains a turn, that segment of the road needs to consist of at least 3 nodes: the node incoming to the turn, the turn node and the node outgoing to the turn. It is a graph such that $R_V$ consists of the locations of each road node, and $R_E$ consists of the connections between road nodes. A road graph can be obtained from OpenStreetMap with software such as the Python library \texttt{osmnx}, and requires specifying an area to obtain the road graph for as an input.

One insight we can make is that the dimensions of the traffic data and the road graph are not necessarily the same; in fact, the size of $R_V$ is often much larger than $T_V$\footnote{We use this assumption throughout the paper; while this might not be the case in traffic settings where there is a large density of traffic sensors, it is unlikely this assumption would be broken in real world due to the sufficiency of fewer traffic sensors for representing traffic flow.}.

\begin{figure}[H]
    \centering
    \includegraphics[width=0.8\linewidth]{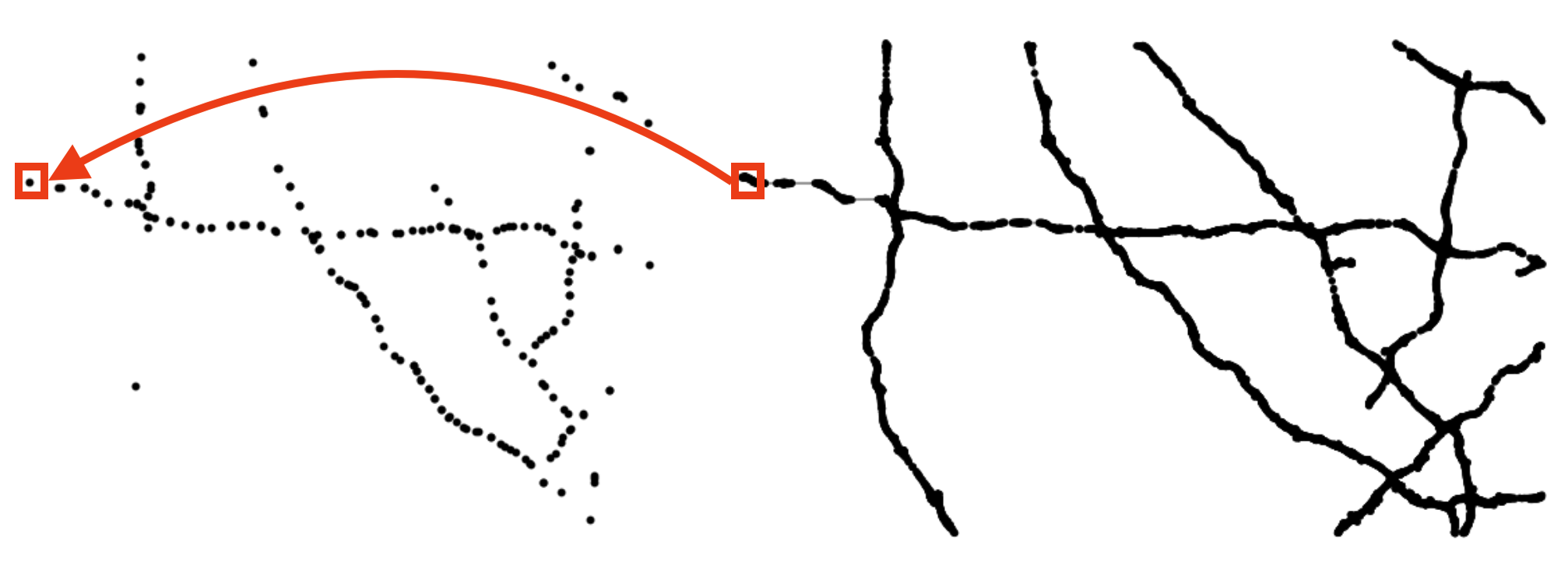}
    \caption{Left: the traffic data $T = (T_V, T_E)$. Right: the road graph $R = (R_V, R_E)$.}
    \label{fig:enter-label}
\end{figure}

Therefore, we need to find some mapping ``down'' from the road graph into the traffic data. One way to do this is to break down $R_V$ such that it is grouped into `clusters' or `blocks' based on $T_V$ in some form. We can use \textit{proximity} as a measure of determining whether a certain traffic nodes $t \in T_V$ belongs to a cluster from $R_V$; more formally, if $\mathrm{pos}(t)$ returns an $x$-$y$ (or latitude-longitude) coordinate pair in $\mathbb R^2$, we wish to compute \[
\{x \in R_V: \| \mathrm{pos}(t) - \mathrm{pos}(x) \| < \varepsilon\}
\]

While this works to identify road nodes with nearby traffic nodes, this does not work as a function because there are scenarios with larger $\varepsilon$ where road nodes are associated with multiple traffic nodes. To solve this problem, we can introduce an equivalence relation $R$ that \textit{partitions} $R_V$ into clusters, and then use $\varepsilon$ as a hyperparameter for pruning OSM nodes that have less geographic similarity with the `root' traffic node.

The equivalence relation is especially useful as it allows us to use a tool from graph theory to construct a graph that maintains the connections of the road graph $R$ while reducing it to the size of the traffic graph $T$. This concept is called a \textit{quotient graph}. Formally, if we have a graph $G = (V, E)$, and some equivalence relation $R$ that partitions $G$ into clusters $C_i$ such that if $v \in C_i$ is adjacent to $v \in C_j$, then $C_i$ is adjacent to $C_j$, then we can define the \textit{quotient graph}
     \[Q = (V/R, \{([u]_R, [v]_R \mid (u, v) \in E\}).\]
Using this information, we define a \textit{traffic quotient graph} as the quotient graph of the traffic data and a road graph as queried by OpenStreetMap. The resultant traffic quotient graph has great similarity with the actual road connections of the road graph. The key difference is that the nodes in the traffic quotient graph are not arbitrary nodes that define direction changes on each road. Instead, the nodes are precisely the nodes from \textit{traffic}. Therefore, the only information stored is the direct road connectivity between nodes, which should help infer the transfer of traffic flow.

\begin{algorithm}
\caption{Constructing a traffic quotient graph}
\begin{itemize}
    \item Load the original traffic sensors, with coordinates.
    \item \textbf{(OSM query)} Retrieve OpenStreetMap data bounded by the coordinates of traffic sensors (either a convex hull or envelope), and define $G = (V, E)$.
    \item Find an equivalence relation $R$ between each set of traffic nodes, by:
    \begin{itemize}
        \item For each traffic sensor, find the nearest road node based on coordinates
        \item For each road node, group into clusters based on the nearest road node and define an equivalence relation $R$ based on this.
    \end{itemize}
    \item Compute the quotient graph $Q = (V/R, \{([u]_R, [v]_R \mid (u, v) \in E\})$.
    \item Prune the graph with $\varepsilon$ if required, such that each node in a cluster is a minimum distance from the root node of the cluster.
\end{itemize}
\end{algorithm}

In the case of the previously discussed METR-LA and PEMS-BAY datasets, the output traffic quotient graphs show great similarity with the original roads. Pruning using an $\varepsilon$ parameter increases the similarity with the true road topology as ``noisy'' nodes are not included in the clusters.

\begin{figure}[H]
    \centering
    \includegraphics[width=0.4\linewidth]{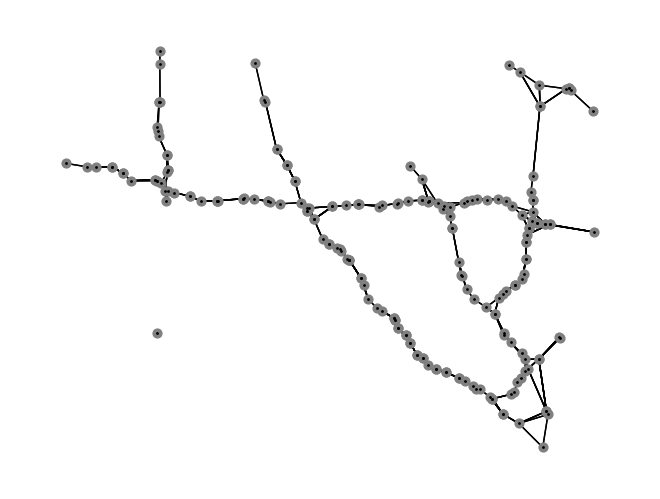}
    \includegraphics[width=0.4\linewidth]{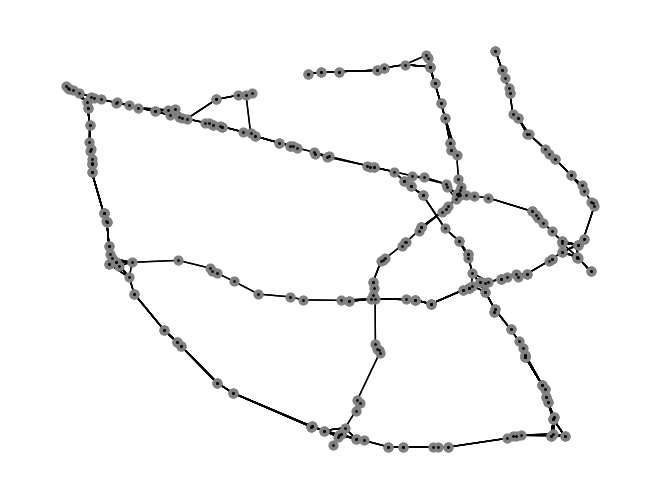}
    \caption{The constructed traffic quotient graphs for two different datasets. Left: METR-LA. Right: PEMS-BAY. Note that for METR-LA the resultant traffic quotient graph has an isolated node, but when using Graph WaveNet connections can still be learned using the self-adaptive adjacency matrix.}
    \label{fig:enter-label}
\end{figure}

\section{OpenStreetMap feature extraction}

While we constructed the OpenStreetMap graph earlier and discussed how we can combine it with the traffic data, the original motivation for the linking of traffic and OSM data was to add relevant information around traffic nodes. Here we discuss the retrieval of this \textit{spatial context}, and its insertion into a \textit{feature matrix}.

A feature matrix $f(Q) \in \mathbb R^{N \times F}$ associated with a traffic graph $Q$ is an $F$-dimensional matrix containing information about every node of the traffic graph $Q$. Note that this feature matrix stores information about the \textit{road nodes} that the traffic nodes are centred on (see the previous section for more details), not the original traffic nodes.

\subsection{The structure of OpenStreetMap data}

When querying a road graph, we obtain both the graph itself as well as two sets of data (in the Python library \texttt{osmnx}, these are called \texttt{GeoDataFrame}s). They are \begin{itemize}
    \item \texttt{gdf\_nodes}, which store information about road nodes, and;
    \item \texttt{gdf\_edges}, which store information about edges connecting road nodes.
\end{itemize}

\texttt{gdf\_nodes} primarily has information on the position (and other node features) of road nodes, while \texttt{gdf\_edges} has information on the speed limits of roads, the length of roads, the type of roads, their one-way status as well as more contextual information such as road name. Importantly, each edge in \texttt{gdf\_edges} is defined as a pair between nodes in \texttt{gdf\_nodes}. We call features that we can extract directly from \texttt{gdf\_nodes} or \texttt{gdf\_edges} \textit{direct features}, and these include:
    \begin{itemize}
    \item \texttt{x}, the $x$-coordinate (longitude) of the node.
    \item \texttt{y}, the $y$-coordinate (latitude) of the node.
    \item \texttt{lanes}, the number of lanes of the edge going into or out of the node.
    \item \texttt{maxspeed}, the maximum speed of the edge going into or out of the node.
    \end{itemize}

    \begin{figure}[H]
        \centering
        \includegraphics[width=0.9\linewidth]{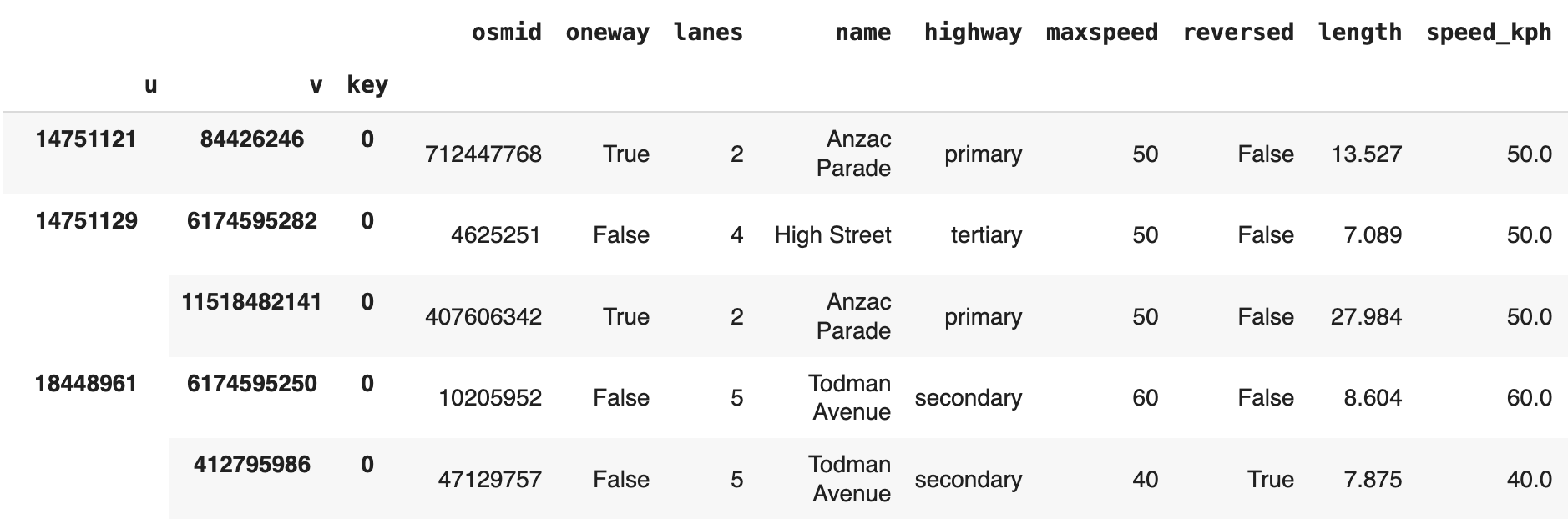}
        \caption{A sample of \texttt{gdf\_edges}, a \texttt{GeoDataFrame} generated around UNSW.}
        \label{fig:enter-label}
    \end{figure}

However, there is also data that is not directly available in either. Additional data, such as the number of points of interest around a road node, needs to be obtained with an additional network call to OpenStreetMap. We call these features \textit{derived features}. In our experiments, we use just one derived feature: \texttt{amenities}, which counts the number of nearby \texttt{amenity} nodes (not road nodes, \textit{any} nodes). This is used ``for describing useful and important facilities for visitors and residents'', and we use this as a proxy feature for the general amenity of an area; the hypothesis is that this can provide some input to the structure of what we would consider a centre of business, entertainment or general activity.

In our experiments, we will perform an ablation study which evaluates the importance of each of these features in improving performance.

\subsection{$u$/$v$ node-edge pairing}

Note that for our method, we wish to associate features with individual nodes, not edges. As seen in the earlier figure, edges are given in the form $(u, v)$, where $u$ and $v$ are nodes; however, we would like to associate edges to \textit{nodes} when the node is either $u$ or $v$. To do this, in our method we take the edges coming in and out of an OSM node and take the maximum value for each feature that relates to edges (assuming any low outlier values come from e.g. on-ramps onto highways). We call this \textit{$u$/$v$ node-edge pairing}. This is just one way to match up edges with nodes; averages, modes and other summary statistics are also possible.

\subsection{Feature matrices and traffic quotient graphs}

In the previous section, we discussed the importance of computing the ``true'' road connectivity between traffic sensors through the construction of the traffic quotient graph. With feature matrix extraction, we are also extracting important data about traffic nodes, but this is data attached to a road node in a graph. However, we can compute the feature matrix for a subgraph \textit{around a node} to get features about a node and its neighbourhood. With this in mind, in the next section we will bring these two types of data together (the graph itself, and the features on each node of the graph), and introduce the concept of a \textit{geometric encoder} and contrastive pre-training.

\section{Contrastive pre-training with the geometric encoder}

\subsection{Motivation and structure}

As discussed earlier with spatially gated addition in the SCPT method, the output of the pre-trained encoder is concatenated with the output of the original prediction model. We wish to output an $F$-dimensional tensor that describes ``geometric information'' about that node.  To build our representation of a traffic node, we can take a subgraph \textit{around a node}, alongside features associated with traffic nodes on the graph, and feed this information into an encoder which we call this a \textit{geometric encoder}.

How do we extract the important features from the defined input? We can use a neural network for this task, in particular combined with layers discussed previously (such as graph convolutional layers). However, given the unsupervised definition of this task, it is not immediately clear how to train this network. We propose a contrastive approach where we generate pairs, in a similar manner to SimCLR \cite{chen2020simpleframeworkcontrastivelearning}.

    \begin{figure}
        \centering
        \includegraphics[width=\textwidth]{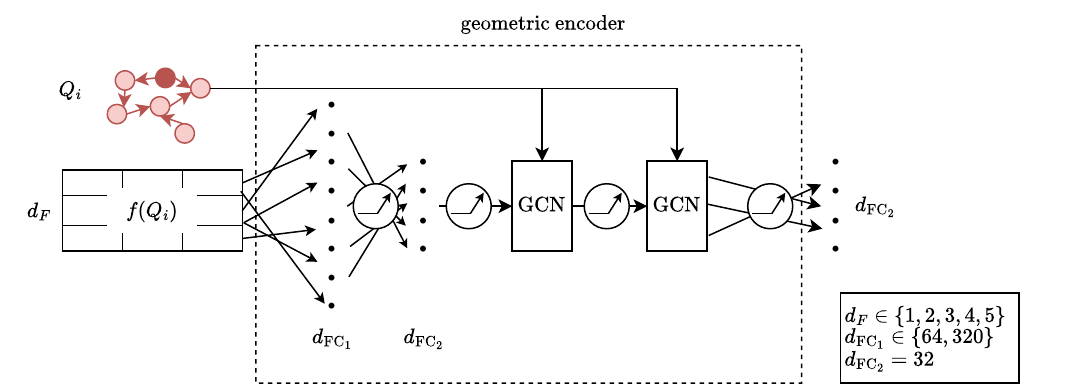}
        \caption{An illustration of the approach. Note we use GraphNorm \cite{cai2021graphnormprincipledapproachaccelerating} after our GCN layers. Our baselines set $d_F = 2, d_{\mathrm{FC}_1} = 64, d_{\mathrm{FC}_2} = 32$ (see the evaluation section for more details).}
        \label{fig:enter-label}
    \end{figure}

The motivation for the neural network architecture is as follows: we use a series of hidden layers to extract hidden relationships from the input feature matrix. We then use a series of two graph convolutional layers which we denote as GCN (although the number of these can be increased) to incorporate information about the graph structure, and learn connections using information propagation. These graph convolutional layers can be followed by GraphNorm layers \cite{cai2021graphnormprincipledapproachaccelerating}, a normalisation method specially designed for GCNs that improve generalisation. Since these are combined with the feature encoder relationships learned from the hidden layers, this gives us a relationship between the two data types\footnote{Implementation note: it is important that the adjacency from the quotient graph and the adjacency from the feature matrix are ordered in the same manner}. ReLU activation units link the layers together before there is a final projection into $d_{\mathrm{FC}_2}$. It is important to note that this output encodes \textit{information about the neighbourhood around a node}.
    
\subsection{Pair generation with cluster node selection}

With images, there are existing augmentation methods, such as cropping, rotating and noising, that are popular for contrastive representation learning. However, for spatio-temporal graphs, or graphs more generally, it is less clear how to augment data. For our method, we focused on one novel method of data augmentation for generating positive pairs (there are other methods of augmentation that can be studied in future work). The novel method makes use of the clusters that we generated earlier from the quotient graph. 

Since each traffic sensor is matched to multiple road nodes, we can randomly select road nodes from the matching. Since we are working with subgraphs, we repeat this procedure for each node in a size $N$ subgraph to end up with two subgraphs with the same approximate true road connectivity. As previously discussed, we can also add an additional radius threshold $\varepsilon$ from the cluster origin to ensure that the selected road nodes are actually near to the traffic nodes.
    \begin{figure}[H]
        \centering
        \includegraphics[width=\textwidth]{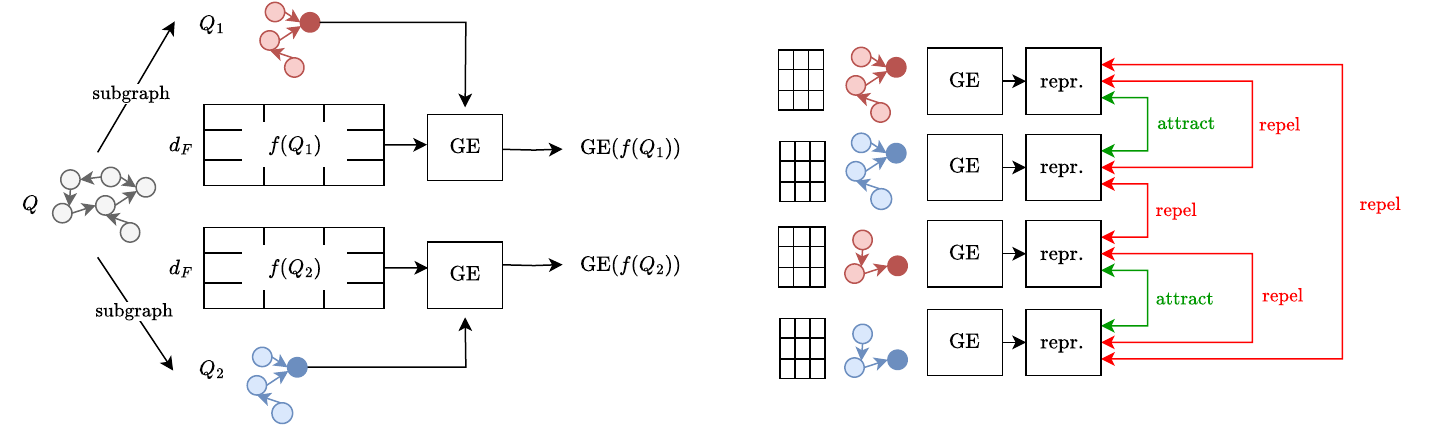}
        \label{fig:enter-label}
        \caption{An illustration of the geometric encoder contrastive training paradigm.}
    \end{figure}
We call these subgraphs $Q_1$ and $Q_2$, and also compute feature matrices (as we will discuss later) called $f(Q_1)$ and $f(Q_2)$ to feed into the geometric encoder. Since these features depend on the road nodes and not the traffic nodes, they vary slightly on each of the nodes, but since they come from similiar geographic positions, the contrastively-trained geometric encoder will still learn similarity within positive pairs. This allows us to learn the concept of ``similar road patterns'', that can be transferred across different parts of the dataset (and ideally, even different datasets).

For example, for the METR-LA dataset, since there are 207 nodes, we choose $N = 64$ (this is a hyper-parameter that can be set). Depending on the input dataset, it might be appropriate to increase or decrease $N$ depending on the density of traffic sensors on the roads.
\begin{figure}[H]
    \centering
    \includegraphics[width=\linewidth]{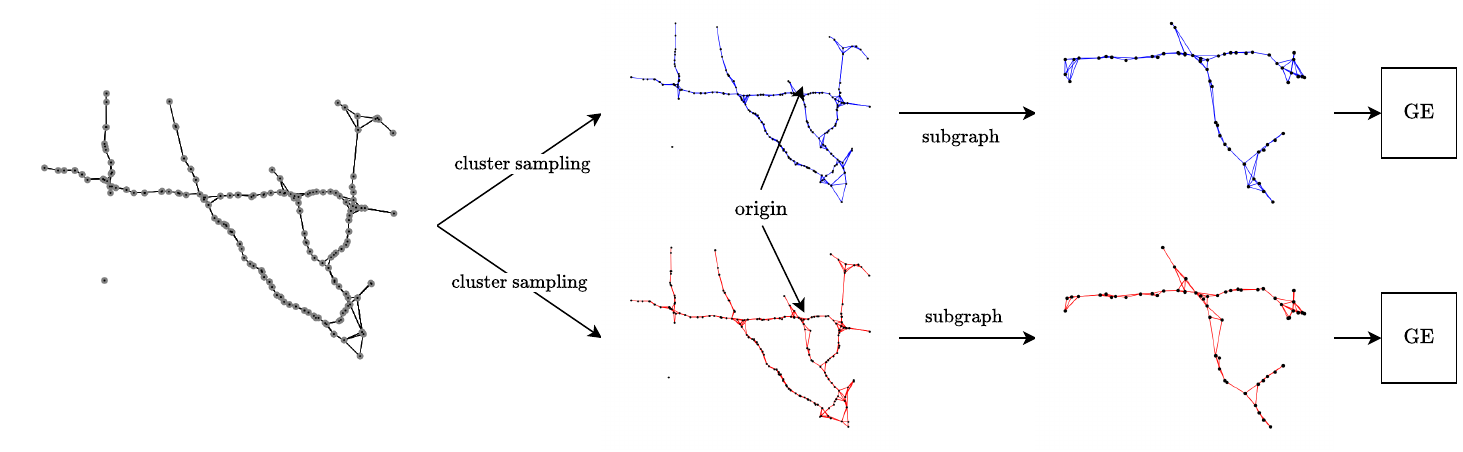}
    \caption{The pair generation process. Notice that we generate two subgraphs with the same origin and feed into the geometric encoder.}
    \label{fig:enter-label}
\end{figure}

In the above figure, inspecting the output subgraphs on the right hand side (before being fed into the geometric encoder) reveals the slight differences in positioning. In fact, since the road node where features are extracted is different, other features such as maximum speed or lanes will also be different, leading to a sort of natural noising. However, the general `topology' of the subgraph is unchanged, so the `gist' of the neighbourhood about the origin is unaffected, which allows the contrastive learning algorithm to still learn similarities between similar neighbourhoods of different nodes.

\begin{algorithm}
\caption{Pre-training pass for an input $x$}
\begin{itemize}
    \item Obtain the quotient graph $Q$.
    \item Derive two sampled quotient graphs $Q_1$ and $Q_2$, where each chooses a random node for each cluster.
    \item Reduce $Q_1$ and $Q_2$ to size $N = 64$ BFS subgraphs based at the origin $x$ (precisely, the cluster corresponding to $x$ in both $Q_1$ and $Q_2$).
    \item Extract features (and scale using \texttt{MinMaxScaler}) corresponding to each, call this $f(Q_1)$ and $f(Q_2)$.
    \item Pass $(f(Q_1), Q_1)$ and $(f(Q_2), Q_2)$ into the contrastive geometric encoder model.
\end{itemize}
\end{algorithm}

This algorithm has mostly linear or polynomial complexity; feature extraction is $O(F\times S)$, where $F$ is number of features and $S$ is the size of the subgraph. For each epoch (pre-train or train), BFS for each node of a random batch size $B$ has to be computed, which is $O(B(V + E))$, where $(V, E)$ come from the subgraph. 

\subsection{The combined pre-trained forecasting model}

Finally, we show the algorithm for training a spatio-temporal forecasting model using this pre-training paradigm. The combination between the pre-training encodings and training weights is done using \textit{spatially gated addition} as discussed earlier.

\begin{algorithm}
\caption{Graph WaveNet, pre-trained with the geometric encoder}
\begin{itemize}
    \item Compute a spatio-temporal split for pre-training and training.
    \item For $E_P = 50$ pre-training epochs:
    \begin{itemize}
        \item Compute a pre-training pass for $B = 64$ nodes, calculating training and validation loss.
        \item If the validation loss is lower, save this as the best encoder weights.
    \end{itemize}
    \item For $E_T = 100$ training epochs:
    \begin{itemize}
        \item Compute the encoder embeddings for each of the nodes in the training epoch.
        \item Run a pass of the Graph WaveNet (or other spatio-temporal forecasting model), integrating the weights of the encoder using spatially-gated addition (SGA).
        \item If the validation loss is lower, save this as the best training weight.
    \end{itemize}
    \item Evaluate the model.
\end{itemize}
\end{algorithm}

\section{Evaluation}

\subsection{Experimentation}

We ran experiments with the Graph WaveNet forecasting model augmented with our geometric encoder with a number of parameter settings. The addition of the geometric encoder encodings was done using spatially gated addition (SGA) as in SCPT \cite{Prabowo_2023}, and therefore the codebase for SCPT was used as a template for performing these experiments.

An ablation and hyperparameter study was conducted with the following variables:
\begin{itemize}
    \item the removal of GraphNorm \cite{cai2021graphnormprincipledapproachaccelerating} layers
    \item the reduction of hidden layer dimension from 320 to 64
    \item learning rate: 0.0001, \textbf{0.0003}\footnote{This was chosen as a baseline as it performed best relative to learning rates.}, 0.001
    \item feature count $F$: 1, \textbf{2}\footnote{The speed and amenities features appeared to be the least noisy in testing.}, 3, 4, 5
    \item cluster radii\footnote{This is the additional threshold radius in latitude/longitude coordinates where nodes from a cluster can be pulled from for contrastive pairs.} $\varepsilon$ : \textbf{0.01}, 0.1
    \item subgraph size $N$: \textbf{64}, full graph
\end{itemize}

All variants were run with 10 seeds and averaged, with 50 pre-epochs and 100 epochs. As is common in machine learning, we follow a 70\%/10\%/20\% train/validation/test split for both our pre-training of the encoder and model training.

\subsection{Results}

Below, we show the results of the different variants on the three different datasets, highlighting both improved training and test set performance.

\begin{table}
    \centering
    \begin{tabular}{l|c|c|c|c|c|c}
        Named variant & $F$ & $\varepsilon$ & $N$ & Hidden & LR & GraphNorm \\\hline
        \texttt{features-1} & 1 & 0.01 & 64 & 320 & 0.0003 & \cmark\\
        \texttt{features-2} & 2 & 0.01 & 64 & 320 & 0.0003 & \cmark\\
        \texttt{features-3} & 3 & 0.01 & 64 & 320 & 0.0003 & \cmark\\
        \texttt{features-5} & 5 & 0.01 & 64 & 320 & 0.0003 & \cmark\\
        \texttt{larger-radius} & 2 & 0.1 & 64 & 320 & 0.0003 & \cmark\\
        \texttt{larger-subgraph} & 2  & 0.01 & $\infty$ & 320 & 0.0003 & \cmark\\
        \texttt{less-hidden} &2 & 0.01 & 64 & 64 & 0.0003 & \cmark\\
        lr-0001 &2 & 0.01 & 64 & 320 & 0.0001 & \cmark\\
        \texttt{lr-0003} &2 & 0.01 & 64 & 320 & 0.0003 & \cmark\\
        \texttt{lr-0010} &2 & 0.01 & 64 & 320 & 0.0010 & \cmark\\
        \texttt{no-graph-norm}&2 & 0.01 & 64 & 320 & 0.0003 & \xmark\\
    \end{tabular}
    \caption{A comparison of the different tested variants of the geometric encoder pre-training paradigm.}
    \label{tab:my_label}
\end{table}

\begin{figure}[H]
    \centering
    \includegraphics[width=0.45\linewidth]{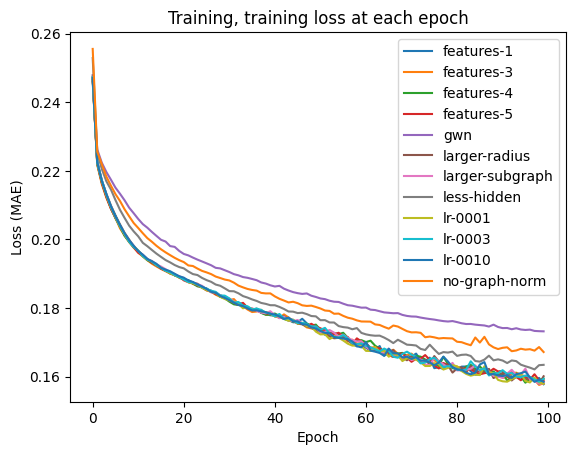} \includegraphics[width=0.45\linewidth]{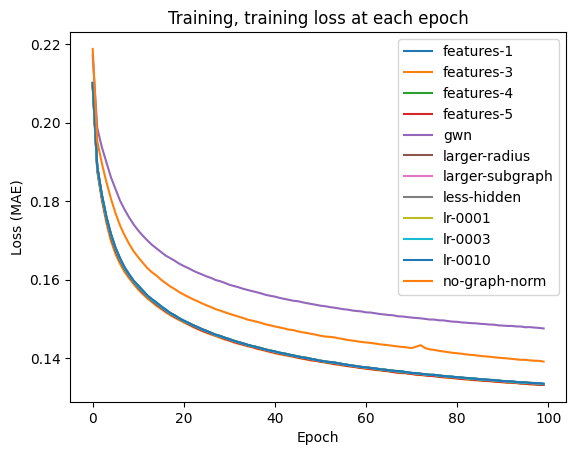}
    \includegraphics[width=0.45\linewidth]{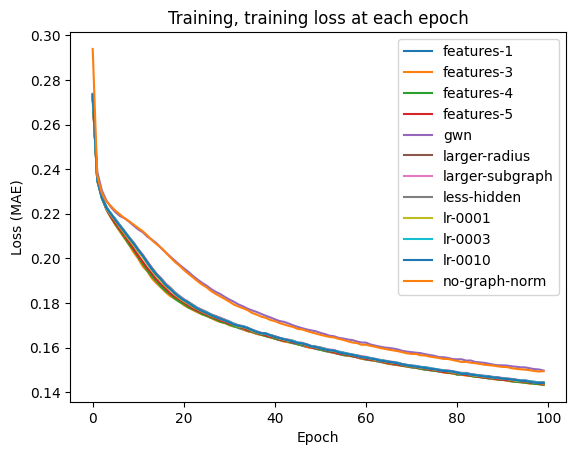}
    
    \caption{Top: training loss on all nodes on METR-LA, PEMS-BAY and PEMS-D7(m), averaged across 10 seeds.}
\end{figure}

Across our experiments, in most cases, we were able to improve performance on the validation set using the geometric encoder approach \textit{relative to Graph WaveNet}, with different combinations of variables in certain scenarios.\footnote{While a full study of hyperparameters using Optuna was considered, the lack of clear variables that improved validation performance makes this sort of study difficult.} The geometric encoder consistently resulted in a faster increase in training loss, but this did not directly translate to validation loss reduction. This could be a signal of potential overfitting, although test set performance is promising for all three datasets. In addition, early epoch loss improvement on training loss is a promising sign that the additional encoder information benefits the training of the underlying forecasting model, and it is possible model adjustments will alleviate any validation overfitting weaknesses.

However, the conclusion from these experiments is that additional non-traffic data, when applied in a pre-training scheme to spatio-temporal graph traffic forecasting, \textit{can} improve model performance, and that there is much potential to experiment with hyperparameters and models in this area.

\begin{table}
    \centering
    \begin{tabular}{l|l|l|l}
        Variant & METR-LA & PEMS-BAY & PEMS-D7(m) \\\hline
        \texttt{features-1} & 12.77 & 6.05 & 8.84\\
        \texttt{features-3} & 12.87 & 6.12 & 9.01\\
        \texttt{features-4} & 12.83 & 6.06 & 8.7\\
        \texttt{features-5} & 12.97 & 6.53 & 8.92\\
        \texttt{gwn} & 12.86 & 6.09 & 9.03\\
        \texttt{larger-radius} & 12.97 & 6.17 & 9.12\\
        \texttt{larger-subgraph} & 12.88 & 6.32 & 9.14\\
        \texttt{less-hidden} & \textbf{12.74} & 6.25 & 9.88\\
        \texttt{lr-0001} & 12.93 & 5.97 & \textbf{8.67}\\
        \texttt{lr-0003} & 12.86& \textbf{5.89} & 9.02\\
        \texttt{lr-0010} & 12.92& 5.94 & 9.31\\
        \texttt{no-graph-norm} & 12.87 & 5.93 & 9.06
    \end{tabular}
    \caption{Mean MAE test set performance for 12-step-ahead traffic prediction on the three datasets.}
    \label{tab:my_label}
\end{table}

\begin{table}
    \centering
    \begin{tabular}{r|l|l|l}
        no. timesteps & METR-LA & PEMS-BAY & PEMS-D7(m)\\\hline
        all steps & \texttt{less-hidden} & \texttt{lr-0003} & \texttt{lr-0001}\\
        1 steps & \texttt{features-3} & \texttt{gwn}& \texttt{lr-0001}\\
        2 steps & \texttt{features-3} & \texttt{lr-0003}& \texttt{lr-0001}\\
        3 steps & \texttt{gwn}& \texttt{lr-0003}& \texttt{lr-0001}\\
        4 steps & \texttt{gwn}& \texttt{lr-0003}& \texttt{lr-0001}\\
        5 steps & \texttt{less-hidden}& \texttt{lr-0003}& \texttt{lr-0001}\\
        6 steps & \texttt{less-hidden}& \texttt{lr-0003}& \texttt{lr-0001}\\
        7 steps & \texttt{less-hidden}& \texttt{lr-0003}& \texttt{lr-0001}\\
        8 steps & \texttt{less-hidden}& \texttt{lr-0003}& \texttt{lr-0001}\\
        9 steps & \texttt{less-hidden}& \texttt{lr-0003}& \texttt{lr-0001}\\
        10 steps & \texttt{less-hidden}& \texttt{lr-0003}& \texttt{lr-0001}\\
        11 steps & \texttt{less-hidden}& \texttt{lr-0003}& \texttt{lr-0001}\\
        12 steps & \texttt{less-hidden}& \texttt{lr-0003}& \texttt{lr-0001}\\
    \end{tabular}
    \caption{The best performing variants (mean MAE) for each timestep prediction for all nodes, for METR-LA,  PEMS-BAY and PEMS-D7(m) respectively. Note that each step is 5 minutes ahead.}
    \label{tab:my_label}
\end{table}

In terms of performance, there was an additional pre-training time per epoch of 2 seconds for $N=64$ subgraphs (this increases proportionally with larger subgraphs, due to feature queries being required on each node in the subgraph). Baseline epoch training times were 17 seconds on METR-LA and 40 seconds on PEMS-BAY. Overall, there was an approximately 50-60\% increase in training time per epoch with pre-training, which could be reduced with more efficient learning schemes or more compact architectures.

\section{Further work}

During this paper we identified a number of potential research directions that were not covered in the work for this paper. In this section, we will discuss some of these directions. Some of these directions are limited to our geometric encoder model, while others are more general (such as different pre-training paradigms or target problems).

\subsection{Generalisation across datasets}

In this paper, we focused on the scenario of pre-training a geometric encoder on a dataset $A$, and training Graph WaveNet (a spatio-temporal forecasting model) on $A$. In our setup, pre-training with the geometric encoder requires a matching between traffic sensors. However, since the pre-training algorithm does not actually use traffic \textit{data}, this can be applied to any reasonably spaced graph that measures connections between nodes without the quotient graph step, even if there are no traffic sensors there. This could be done in a generative manner, and in a way would allow us to generate synthetic traffic data, although the details of this approach still need to be elaborated upon.

\subsection{Model refinements}

In particular, in the SimCLR paper, there is discussion about how the sum of different augmentation techniques produced better results than individual augmentations. While our cluster selection approach provides some augmentations, additional augmentations should be explored for spatio-temporal data. The augmentations used in SimCLR included cropping, resizing, color distortion, rotation, cutout, and noising; while these might not transfer as easily in a graph context, some such as \textit{flipping} should do so. GraphCL \cite{You2020GraphCL} discusses methods for graph augmentations, such as node dropping and edge perturbation, which may be appropriate in a spatio-temporal context as well (assuming that perturbations do not result in misinterpretation of geographical features).

\subsection{Traffic state estimation}

While we have focused on the task of predicting future traffic given past traffic, there is a more generalised problem that warrants further investigation; the \textit{traffic state estimation} problem. Traffic state estimation is the `completion' or estimation of unobserved traffic variables (e.g. speed) given incomplete data, which can be useful for traffic management purposes \cite{Shi_Mo_Di_2021}. There are existing papers that apply (for example physics-based approaches for traffic state estimation, but there is potential in applying geometric encoder information and `road features' to this problem given the correlation between amenities and traffic-related events. For example, areas with higher entertainment amenity count might attract higher traffic counts in rare scenarios during concerts or other events.

\subsection{Better urban features}

We studied the \texttt{amenities} feature, which gives us an approximate indication of the number of POIs within a radius of a point. However, there is more detailed information that we can gather around a traffic node, especially since amenities also include information about the type of amenity. We could compute some sort of \textit{amenity score}, which weights amenities higher if they are closer, or filter amenities based by the type of amenity and the time of day that the amenity is primarily associated to (for example, office amenities might be more relevant for predicting traffic in peak hours). We could even compute other features that are not realted to amenities, such as general street structure around the point, or other contextual urban information. All of this information can now be associated with traffic sensors in a meaningful way to improve prediction performance.

\section{Conclusion}

In this paper, we proposed a novel method for constructing traffic quotient graphs that better represent the true connectivity of traffic sensors with context to their road topologies, and proposed a geometric encoder inspired by the contrastive representation learning paradigm that combines information about traffic, roads and road connectivity. We demonstrated that this scheme of pre-training can lead to improvement in traffic forecasting performance across nodes at zero additional dataset collection cost, requiring only a query to existing, well-collected information about the placement of roads and points of interest via OpenStreetMap.

\bibliography{sn-bibliography} 


\end{document}